\documentclass{article}

\PassOptionsToPackage{numbers}{natbib}

\usepackage[preprint]{neurips_2025}

\usepackage[utf8]{inputenc} 
\usepackage[T1]{fontenc}    
\usepackage{hyperref}       
\usepackage{xurl}            
\usepackage{doi}
\usepackage{booktabs}       
\usepackage{amsfonts}       
\usepackage{nicefrac}       
\usepackage{microtype}      
\usepackage{xcolor}         
\usepackage{todonotes}
\usepackage{wrapfig}
\usepackage{csquotes}
\usepackage{comment}
\usepackage{tcolorbox}

\title{AI for Scientific Discovery is a Social Problem}

\author{%
  Georgia Channing$^{\dagger}$\thanks{Equal Contribution. $\dagger$ University of Oxford} \\
  Hugging Face \\
  \texttt{georgia@huggingface.co}
  \And
  Avijit Ghosh\footnotemark[1] \\
  Hugging Face \\
  \texttt{avijit@huggingface.co}
}

\begin{document}

\maketitle


\begin{abstract}
Artificial intelligence (AI) is being increasingly applied to scientific research, but its benefits remain unevenly distributed across different communities and disciplines.
While technical challenges such as limited data, fragmented standards, and unequal access to computational resources are already well known, social and institutional factors are often the primary constraints.
 Narratives emphasizing autonomous ``AI scientists,'' the underrecognition of data and infrastructure work, misaligned incentives, and gaps between domain experts and machine learning researchers all limit the impact of AI on scientific discovery. Four interconnected challenges are highlighted in this paper: community coordination, the misalignment of research priorities with upstream needs, data fragmentation, and infrastructure inequities. We argue that addressing these challenges requires not only technical innovations but also intentional community-building efforts, cross-disciplinary education, shared benchmarks, and accessible infrastructure. We call for reframing AI for science as a collective social project, where sustainable collaboration and equitable participation are treated as prerequisites for achieving technical progress.
\end{abstract}

\section*{Introduction}

Artificial intelligence (AI) is being increasingly applied to scientific research, although the extent, timescale, and scope of this contribution remain open questions.
Recent successes have demonstrated the transformative potential of AI across multiple domains. The protein structure prediction method of AlphaFold~\cite{jumper2021alphafold}, which was recognized with the 2024 Nobel Prize in Chemistry~\cite{deepmind2024nobel}, has been extended by AlphaFold3 to predict interactions between proteins, DNA, RNA, and small molecules with improvements of at least 50\% over the existing methods~\cite{abramson2024alphafold3}. In materials science, Google DeepMind's GNoME has discovered 2.2 million new crystal structures, including 52,000 novel lithium-ion conductors, with external researchers already synthesizing 736 of these predictions~\cite{merchant2023gnome,szymkuc2024insilico}. Drug discovery has seen similar advances. Insilico Medicine's ISM001-055 became the first fully AI-designed drug to reach phase II trials with positive results; it was developed in under 18 months at approximately 10\% of the traditional costs~\cite{insilico2024ism001}. Additionally, BenevolentAI used AI to repurpose baricitinib for addressing COVID-19, receiving FDA authorization within a year~\cite{benevolentai2020baricitinib}. Climate modeling has advanced with AI systems such as Google's NeuralGCM matching the 10-15-day forecasts of traditional models with enormous computational savings~\cite{bi2023neuralgcm,lam2024graphcast}, NVIDIA's StormCast superresolving atmospheric data 1,000x faster using 3,000x less energy~\cite{nvidia2024stormcast}, and NASA-IBM's Prithvi Weather-Climate foundation model improving the resolution of regional climate modeling ~\cite{nasa2024prithvi}. However, these benefits have thus far been concentrated in domains with strong data infrastructure and well-coordinated research communities~\cite{horton2025materialsproject}.

For the purposes of this paper, we define \textit{scientific discovery} as the process of generating new explanatory understanding of natural phenomena through systematic investigation, experimental validation, and integration with existing theoretical frameworks~\cite{de_Regt_2020}. In some domains, particularly physics and chemistry, this explanatory work is often articulated through mechanistic models that identify causal relationships and structural principles. In other domains, explanation may take different forms while serving the same epistemic function: enabling researchers to make sense of phenomena, anticipate novel observations, and guide experimental design. Our concern is not with mechanisms per se, but with whether AI systems support the kinds of explanatory practices through which scientific communities develop shared understanding. 

This contrasts with merely achieving predictive accuracy: a model that correctly forecasts the binding affinity of a protein but provides no insights into which molecular interactions drive that binding represents a technical achievement but not necessarily a scientific discovery. We therefore define success in AI for science along multiple dimensions simultaneously. Successful applications demonstrate strong predictive performance on well-defined tasks, but they also yield explanatory insights that reveal causal relationships and generalizable principles. They enable experimental validations that confirm computational predictions in physical systems and integrate with domain knowledge to advance theoretical understanding. Critically, they maintain broad accessibility, enabling researchers across different institutions and geographies to build upon advances~\cite{krenn2022scientific}. Systems that merely optimize prediction accuracy on benchmark datasets, however technically impressive, fall short of this standard.

AI applications in science take several forms, each with distinct technical requirements but common social barriers. Some approaches focus on accelerating existing workflows: predictive modeling like AlphaFold forecasts protein structures~\cite{jumper2021alphafold}, simulation surrogates enable molecular dynamics at previously impossible scales~\cite{Kochkov_2021,Fiedler_2022}, and laboratory automation combines AI with robotics to speed experimental cycles~\cite{szymkuc2024insilico}. Others tackle search and synthesis problems: design algorithms explore chemical spaces for materials and drugs~\cite{merchant2023gnome,insilico2024ism001}, while large language models mine literature to suggest research directions~\cite{nordmann2025evaluating, qureshi2023are, zhou2024hypothesis}. Increasingly, researchers are developing foundation models for science: large-scale pretrained systems designed to capture general principles across chemistry, biology, and physics, then fine-tune for specific tasks with limited data~\cite{neurips2024foundation,simons2024polymathic}. Despite their technical diversity, these applications share fundamental challenges rooted in how scientific communities organize, fund, and value different types of contributions.

\begin{figure}
    \centering
  \includegraphics[width=0.49\textwidth]{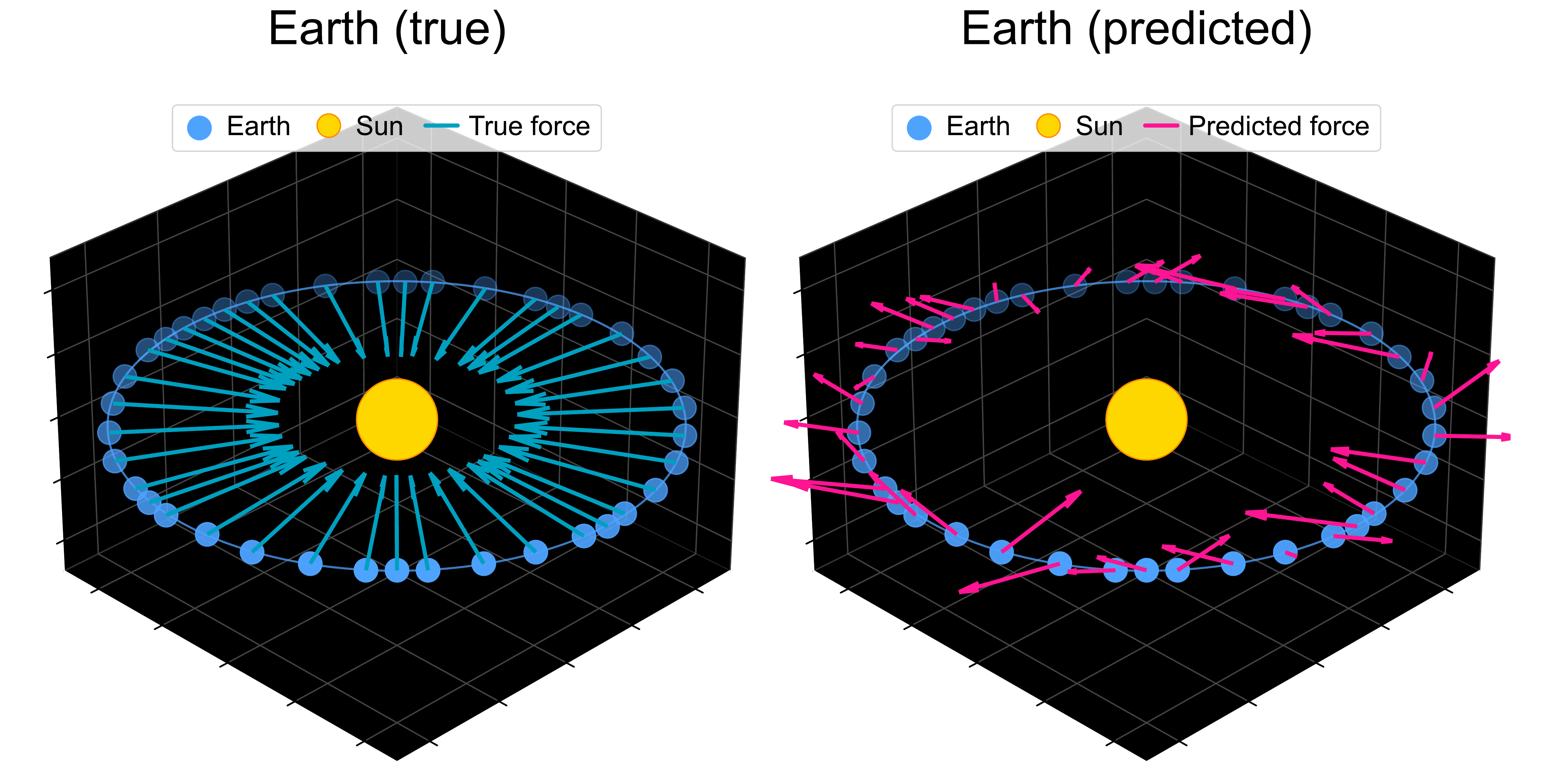}
    \caption{\small \textbf{Predictive accuracy without mechanistic understanding.} This figure from~\cite{vafa2025foundationmodelfoundusing} contrasts the true Newtonian forces (left) and the predicted forces (right) learned by a transformer-based foundation model with high accuracy when predicting planetary trajectories. Although the model performs well on the task it was fine-tuned for, it has not learned an inductive bias toward true Newtonian mechanics. General-purpose models do not necessarily aid in specific scientific understanding cases.}
  \label{fig:sun}
\end{figure}

A critical distinction exists between predictive performance and scientific understanding. As illustrated in \autoref{fig:sun}, a transformer-based model can achieve high accuracy when predicting planetary trajectories without learning the underlying Newtonian mechanics~\cite{vafa2025foundationmodelfoundusing}. Current commercial AI development disproportionately emphasizes literature synthesis over experimental validation and mechanistic investigation~\cite{aisnakeoil_ai_slow_science}, risking the conflation of information retrieval with genuine discovery~\cite{krenn2022scientific}.


Four critical barriers that are preventing the democratization of AI in science are examined in this paper: community dysfunction, which undermines collaboration and perpetuates harmful narratives; misaligned research priorities, which target narrow applications over upstream computational bottlenecks; data fragmentation caused by incompatible standards and social practices~\cite{gisselbaek2025bridging}; and infrastructure inequities, which concentrate power within privileged institutions~\cite{ahmed2020dedemocratization}. While these issues manifest as technical challenges, their root causes are fundamentally social and institutional: undervalued data contributions, educational gaps preventing cross-disciplinary engagement, and the absence of community consensus regarding shared priorities. We then propose the following solutions: strengthening cross-disciplinary collaboration and education, structuring upstream challenges through shared benchmarks, standardizing scientific data practices, and building a sustainable community-owned infrastructure.

Although we distinguish social barriers from technical barriers for clarity, the two are inseparable in practice. Every technical system embeds social assumptions about the data collection process, expertise, labor, incentives, and institutional constraints. The barriers that we describe are therefore interdependent rather than separate categories. We conclude that democratizing AI for science requires treating it as a collective social project where equitable participation is a prerequisite for achieving technical progress.

\section*{Barriers to Scientific Achievement}

\subsection*{Barrier One: Community Dysfunction}

The dysfunction within the AI for science community manifests through harmful narratives, misaligned incentives, and communication breakdowns that prevent effective collaboration. These issues reflect in-depth cultural issues about how the community values different types of contributions, frames the purpose of AI in science, and structures interactions between disciplines.

\paragraph{The AI Scientist Myth}

Recent commercial efforts by OpenAI, Anthropic, and Google DeepMind increasingly emphasize "AI scientists" capable of autonomous research~\cite{lu2024aiscientist, altman2025openai_researcher, anthropic2025lifesciences}. OpenAI announced plans to develop an "intern-level research assistant" by September 2026 and a fully autonomous "legitimate AI researcher" by March 2028~\cite{altman2025openai_researcher}, while Anthropic launched Claude for Life Sciences to "support the entire process, from early discovery through to translation and commercialization"~\cite{anthropic2025lifesciences}. These systems predominantly focus on literature search, summarization, and hypothesis generation from existing text (tasks where large language models demonstrate clear competence) while largely avoiding the experimental validation, iterative refinement, and mechanistic investigation that define scientific discovery.

This emphasis on literature synthesis creates several problems. It devalues experimental and computational research that generates new data rather than synthesizing existing publications. Then, it creates unrealistic expectations about AI's current capabilities, leading to misallocation of resources toward tools that excel at information retrieval while neglecting those that could accelerate actual discovery~\cite{aisnakeoil_ai_slow_science}. Systems optimized for literature synthesis demonstrate serious failure modes when applied to research tasks, including inappropriate benchmark selection, data leakage, metric misuse, and post-hoc selection bias~\cite{luo2025automateseehiddenpitfalls, zhu2025aiscientistsfailstrong}.

The framing of autonomous "AI scientists" also perpetuates problematic attribution patterns. Historians and philosophers of science document how narratives of lone genius systematically erase the contributions of assistants, technical staff, and collaborative teams. These teams often include women and other marginalized groups whose labor proves essential but remains uncredited~\cite{galison2003collective}. For instance, the case of Rosalind Franklin's contributions to the determination of the structure of DNA exemplifies this pattern~\cite{charney2003lone}. When media headlines proclaim "AI scientist discovers X," they reproduce this problematic attribution by granting sole agency to the algorithm while obscuring the data curation, experimental validation, infrastructure development, and domain expertise that made the discovery possible. This erasure has material consequences: it devalues precisely the types of contributions (dataset construction, benchmark development, infrastructure maintenance) critical for democratizing AI in science.

Moreover, scientific creativity fundamentally depends on recognizing which problems are worth investigating. Callon describes this as the "struggles and negotiations" that define research agendas~\cite{callon1980struggles}. Current AI systems excel at optimization within well-defined problem spaces but lack capacity for problem formulation itself: identifying which questions matter, recognizing anomalous results that challenge existing frameworks, and judging when computational predictions warrant experimental follow-up. Framing AI as an autonomous scientist obscures this distinction between search execution and the creative, socially-embedded work of defining what to search for.

\paragraph{Scientific Success Versus AI Success}

A fundamental conflict exists between scientific discovery and AI optimization. As defined earlier, scientific success requires mechanistic insights: understanding \textit{why} a prediction works, which variables matter, and how findings generalize beyond the training conditions~\cite{de_Regt_2020,krenn2022scientific}. This differs from AI success, which prioritizes predictive accuracy, optimization performance, and scaling behavior. A model that accurately predicts drug toxicity levels without identifying the causal biochemical pathways may succeed as an AI system but fail as a scientific tool if it cannot guide rational design processes or explain unexpected failures. Similarly, a material discovery model that recommends promising compounds without explaining their electronic structures provides limited scientific value despite its high predictive performance. These divergences matter because they determine what questions researchers can answer, what experiments they will design next, and whether the findings contribute to the cumulative theoretical knowledge or remain isolated empirical observations.

\paragraph{Rewarding Long-Lasting Impacts}

Another counterproductive tendency within the current research ecosystem is the undervaluing of contributions to data and infrastructure in hiring, publicity, and tenure evaluations~\cite{fecher2015reputationeconomyresultsempirical}. High-quality datasets, particularly those with well-curated metadata, often have far greater long-term impacts than individual model contributions do. Most models are rapidly superseded by marginally improved variants, and their influences diminish within a short time frame. In contrast, datasets and infrastructure frequently underpin entire lines of research and remain useful for decades~\cite{pmid24109559}. The effective half-life of data is therefore likely to be orders of magnitude longer than that of models. Recognizing and rewarding these contributions is essential for building a sustainable foundation for achieving scientific progress with machine learning.

In practice, scientific problems tend to be good candidates for AI
methods when they share several characteristics: high-throughput or
systematically generated datasets; clearly defined evaluation metrics;
well-characterized sources of uncertainty; repeatable experimental
pipelines; and the existence of simulation models that can be used
for data augmentation or rapid approximation purposes. Problems lacking these
features often require substantial infrastructural or organizational
investment before AI methods can be meaningfully applied.

\paragraph{Technical Communication Gaps}

Collaboration failures often stem from fundamental differences among how various fields approach problem formulation and validation work~\cite{Krishnan2025}. Domain scientists prioritize mechanistic understanding and experimental validation, whereas ML researchers focus on predictive performance and computational efficiency. For instance, a biologist studying protein folding cares about whether a model identifies which amino acid interactions drive stability, enabling the design of rational mutations, while an ML researcher may focus on whether the model achieves state-of-the-art accuracy on a benchmark regardless of its interpretability. In climate science, a researcher needs models that respect physical constraints such as the conservation of mass and energy, whereas an ML practitioner might propose a purely data-driven approach that achieves better short-term predictions but fails catastrophically under novel conditions. Different fields may also rely on different tools and software practices, which further reinforces disciplinary boundaries. These different objectives lead to miscommunication about success criteria, acceptable tradeoffs, and validation approaches.

The terminology gap exacerbates these issues. Domain scientists may describe problems using field-specific concepts that do not map directly to standard ML problem formulations. ML researchers may propose solutions that seem promising from a computational perspective but violate the fundamental constraints or assumptions of the target domain. For example, a machine learning model might achieve excellent accuracy in terms of predicting rainfall patterns but do so by overfitting to seasonal cycles in the training data. Without incorporating physical constraints derived from climate dynamics, these predictions fail under shifting climate regimes and provide little scientific insight. When the degree of domain specificity is strong, the complex interdependencies among the methods, epistemic principles, technological practices, and tacit knowledge in a field make it extremely difficult for outsiders to understand how the system operates or the rationale behind practitioners' decisions~\cite{davenport2024interdisciplinary}. However, the impact of large language models on collaboration remains contested. While some evidence suggests that accessible AI tools may lower barriers to entry~\cite{li2024llmcollaboration}, recent research indicates that LLM use may actually reduce collaboration breadth and topical diversity~\cite{hao2026artificial}.

\paragraph{Insufficient Educational Materials}

A critical challenge lies in the lack of educational resources that are tailored for interdisciplinary research. Machine learning practitioners seeking to apply methods in biology, chemistry, and physics lack primers designed for technically mature but domain-naive audiences, whereas domain scientists frequently lack training in computational and ML methods. This bidirectional gap reflects the fact that most researchers are trained in traditional sciences without computing requirements, often leading them to reinvent existing solutions~\cite{scoles2017modern}. Just as mathematics and statistics are fundamental to scientific training, machine learning should now be treated as essential for domain scientists. Without this foundation, collaborative efforts suffer as each group struggles to recognize what the other can contribute~\cite{bianchini2024driversbarriersaiadoption}.

\subsection*{Barrier Two: Misaligned Research Priorities}

The scientific community's focus on narrow, domain-specific applications over upstream computational challenges reveals how academic reward systems shape research agendas. Publication pressure, grant cycles, and disciplinary boundaries create powerful incentives that fragment efforts across hundreds of domain-specific problems rather than mobilizing collective action around high-impact computational bottlenecks.

\begin{table}[h]
\centering
\small
\begin{tabular}{p{0.25\textwidth}p{0.28\textwidth}p{0.38\textwidth}}
\toprule
\textbf{Challenge} & \textbf{Domains Affected} & \textbf{Current Limitation} \\
\midrule
PDE Neural Operators & Weather, plasma fusion, aerodynamics & Regime extrapolation fails (laminar→turbulent, present→future climate) \\
\midrule
High-D Inverse Problems & Materials, drugs, astrophysics & Special-case solutions; no general probabilistic framework \\
\midrule
Rare Event Sampling & Protein folding, extreme weather, reaction barriers & Standard training misses low-probability tails \\
\midrule
Multiscale Coupling & Tissue mechanics, virtual cells & Single-scale models lack mechanical/spatial constraints \\
\midrule
Quantum Many-Body & Superconductors, catalysts & Exponential scaling; sign problem limits accuracy \\
\bottomrule
\end{tabular}
\vspace{0.2cm}
\caption{\textbf{Representative upstream computational challenges with broad scientific impact.} Success in these areas would unlock progress across multiple downstream applications.}
\label{tab:challenges}
\end{table}

\paragraph{Identifying Upstream Computational Bottlenecks}

Certain computational challenges appear across numerous scientific domains, representing high-leverage targets for AI development. Fast density functional theory (DFT) calculations underpin material discovery, catalysis research, and solid-state physics~\cite{Fiedler_2022}. Efficient partial differential equation (PDE) solvers enable real-time fluid dynamics, heat transfer modeling work, and climate simulations~\cite{Kochkov_2021}. Long-timescale molecular dynamics capture biochemical processes and material aging across chemistry, biology, and engineering.

The critical assessment of protein structure prediction (CASP) illustrates how structuring an upstream challenge can catalyze broad progress. By framing protein structure prediction as a recurring community benchmark with standardized evaluation metrics, the CASP identified the key computational bottleneck and created a venue for achieving sustained methodological innovations~\cite{moult1995caspexperiment}. This groundwork ultimately enabled breakthroughs such as AlphaFold, but the critical contribution was the clear problem formulation and consistent evaluation regime that mobilized an entire research community~\cite{jumper2021alphafold}. By focusing collective efforts on a shared upstream challenge, the CASP multiplied the downstream impact relative to that of piecemeal advances in narrow protein engineering tasks.





\paragraph{Examples of High-Impact Computational Problems}

Several computational challenges represent high-leverage targets for AI development (\autoref{tab:challenges}). Physics foundation models that learn operator mappings rather than individual solutions could transform weather forecasting, plasma fusion modeling, and aerodynamics, but achieving robust generalization across different regimes (laminar to turbulent flows, present to future climate states) remains an open problem~\cite{Kochkov_2021, bi2023neuralgcm, simons2024polymathic}. The inverse problem of recovering structures from desired properties is central to materials science, drug discovery, and astrophysics, yet current approaches solve only special cases without providing general probabilistic frameworks~\cite{wang2021deep, merchant2023gnome}. Rare event sampling in high dimensions would accelerate discoveries in stochastic systems from protein folding to extreme weather, but standard training procedures focus on average performance and miss low-probability events critical for scientific understanding~\cite{binbin2025sequential, noe2019boltzmann}.

The multiscale coupling problem exemplifies challenges in biological systems spanning angstroms to millimeters. Current virtual-cell initiatives often rely on single-cell transcriptomic models lacking mechanical or spatial constraints~\cite{karr2012wholecell, bunne2024virtualcell}. Computing strongly correlated quantum systems remains limited by exponential scaling and the fermion sign problem, preventing precision engineering of high-temperature superconductors and advanced catalysts~\cite{Rossi2008}. Progress on these upstream challenges would multiply downstream impact compared to fragmented efforts on narrow applications.

\begin{figure*}[t]
\centering
\includegraphics[width=0.9\textwidth]{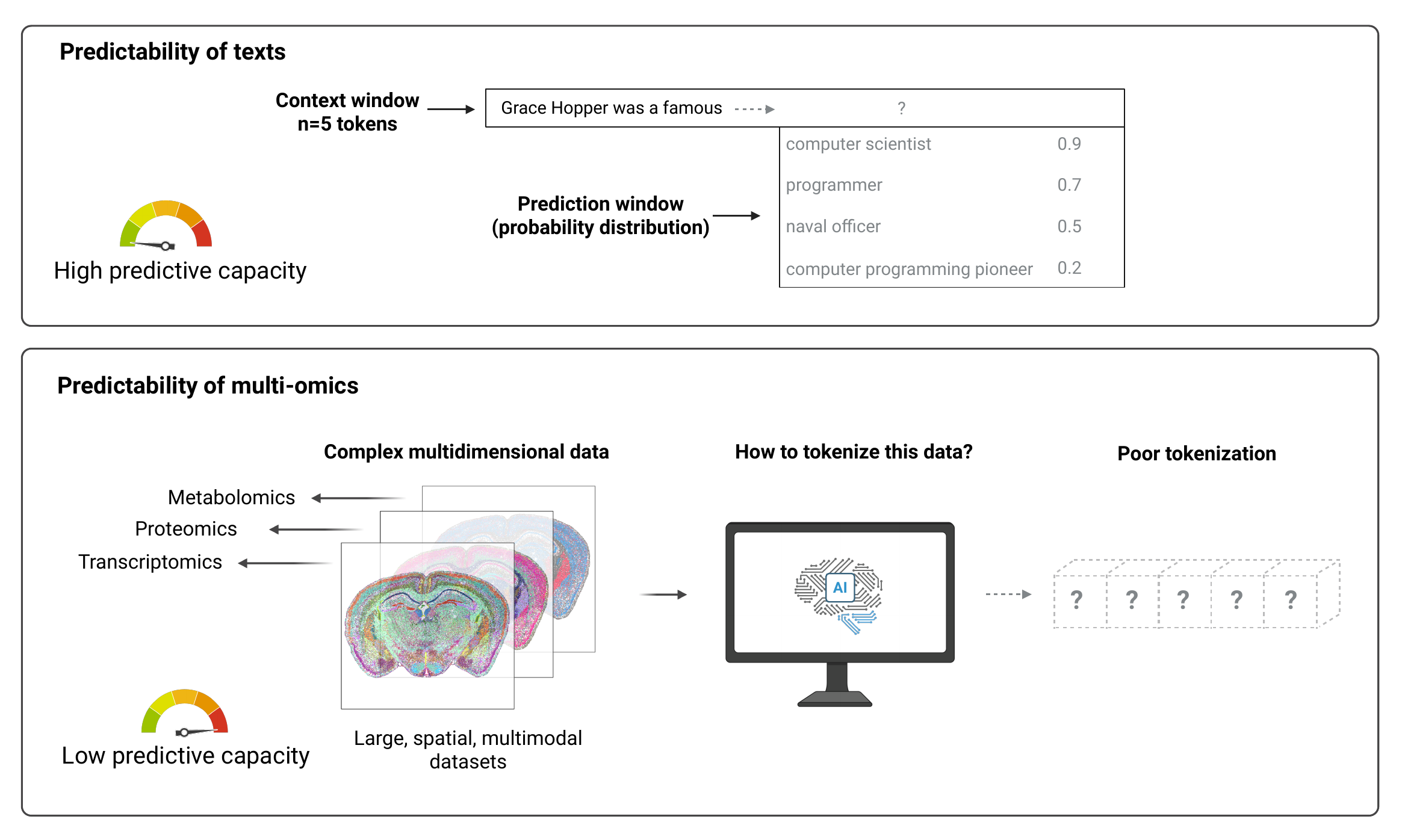}
\caption{\textbf{Scientific data tokenization challenges.} Current AI architectures excel at predicting text tokens (top) but struggle with complex scientific datasets such as multiomics data (bottom), which lack clear tokenization strategies and exhibit low predictive capacities despite their richness and scale.}
\label{fig:multiomics}
\end{figure*}

\subsection*{Barrier Three: Data Challenges}

Data fragmentation in science refers to the scattering of research data across disconnected sources, where it is stored in incompatible formats without shared standards or effective indexing. This makes the discovery, integration, and reuse of such data difficult. The problem stems from a community that has failed to prioritize interoperability and sharing despite generating vast quantities of valuable datasets. Researchers hoard data in proprietary formats, funding agencies do not require standardization, and institutions provide no career rewards for the tedious work of data curation and harmonization. The result is a landscape where most scientific datasets remain locked in incompatible silos, while the high-dimensional, low-sample-size nature of scientific data makes these coordination failures even more costly for AI development work.

\paragraph{The High-Dimensional, Low-Sample-Size Problem}

Consider protein structure prediction. While AlphaFold succeeded with millions of protein sequences, most scientific domains lack equivalent data abundance conditions~\cite{jumper2021alphafold}. A typical cancer genomics study might have 10,000 features (genes) but only 100 patients. Whole-slide imaging (WSI) involves billions of pixels with complex spatial hierarchies, yet training sets rarely exceed thousands of samples~\cite{camelyon}. The current transformer architectures, which were designed for text where small context windows suffice, generally fail to capture the extended spatial relationships that are essential for achieving scientific understanding~\cite{chen2022scalingvisiontransformersgigapixel}. As illustrated in \autoref{fig:multiomics}, while AI models excel at predicting text through established tokenization strategies, scientific datasets such as multiomics datasets, including transcriptomics, proteomics, and metabolomics, contain complex spatial, temporal, and multimodal relationships that resist straightforward tokenization approaches \cite{arnett2024llama_tokenizer}, resulting in poor predictive performance despite the large scale and richness of the underlying data.

The context window limitation becomes stark in biological systems. A transformer trained on protein sequences might capture local amino acid patterns \cite{dotan2024effect}, but understanding protein functions requires considering interactions across the entire three-dimensional structure--often requiring 1000x-larger context windows than the current models support. Random autoregressive rollout works for text generation because the local context provides meaningful signals, but protein folding depends on long-range interactions that sequence models are inherently limited in their ability to represent; specific architecture modifications are required \cite{sidorenko2024precious2gpt}.

\paragraph{Data Fragmentation and Standardization Challenges}

Beyond architectural limitations, scientific data suffer from severe fragmentation that prevents the development of collaborative models. Research organizations typically manage more than 100 distinct data sources, with 30\% handling more than 1,000 sources~\cite{Pressley2021}. For instance, a 2020 survey revealed that data scientists spend approximately 45\% of their time on data preparation tasks, including loading and cleaning data~\cite{woodie2020_dataprep_time}. Domain-specific formats, inconsistent metadata standards, and incompatible experimental protocols create artificial barriers that fragment knowledge and limit collaborative development. Recent research on machine learning data practices has revealed fundamental challenges: ML researchers struggle to apply standard data curation principles because of the presence of shared terms with nonshared meanings between fields, high interpretative flexibility in terms of adapting concepts, the depth of data curation expertise needed, and the unclear scopes of documentation responsibilities~\cite{bhardwaj2024mlcuration}.

The cancer research field exemplifies these challenges. Harmonizing just nine Cancer Imaging Archive files required 329.5 hours over six months, resulting in the identification of only 41 overlapping fields across three or more files~\cite{basu2019call}. This labor-intensive process must be repeated for each new research initiative, creating insurmountable barriers for smaller institutions that lack dedicated data curation resources.

Similar problems arise outside the life sciences. In astroparticle physics, experiments such as KASCADE and TAIGA produced distributed datasets in heterogeneous, experiment-specific formats, such as ROOT trees, custom binaries, and ASCII logs, that required specialized software just to parse~\cite{haungs2019germanrussianastroparticledatalife}. Because these files were large and monolithic, extracting subsets (e.g., high-energy showers) from them often meant loading gigabytes of irrelevant data. The data were also scattered across institutions with limited metadata, creating significant computational and logistical overhead before a scientific analysis could even begin~\cite{kryukov2019distributeddatastoragemodern}.

\paragraph{Data Format and Evaluation Incompatibilities}

Domain scientists typically generate datasets that are optimized for human interpretation rather than for training machine learning models. Experimental data often come in specialized formats with implicit preprocessing, normalization, and quality control assumptions that ML researchers must reverse-engineer. Different fields use incompatible evaluation metrics, making it difficult to assess whether ML approaches actually improve upon domain-specific baselines.

For example, computational biology datasets often require complex preprocessing pipelines to handle batch effects, missing values, and experimental artifacts \cite{lutge2021cellmixs}. Without domain expertise, ML researchers may train models on processed data without understanding the underlying experimental limitations, leading to overfitting to technical artifacts rather than biological signals.

\subsection*{Barrier Four: Infrastructure Inequity}

The infrastructure inequity encountered in scientific computing reflects broader resource concentration and institutional privilege patterns that go beyond technical necessity. The social dynamics of resource allocation, namely, who is given priority, how decisions are made, and which institutions are considered ``deserving'', ultimately determine whether AI tools become democratized or further centralize scientific capabilities. The challenge lies in building systems that work effectively across different resource constraints while maintaining high scientific quality and enabling collaborative development.

\paragraph{Computational Resource Disparities}

Academic researchers face substantial barriers to accessing computational resources that are adequate for competitively developing AI models. A survey revealed that 66\% of scientists rated their computing satisfaction levels as three out of five or less, with only 10\% having access to state-of-the-art hardware~\cite{khandelwal2025, kudiabor2024ais}. Wait times typically extend from days to weeks, whereas complex application processes favor institutions with existing technical infrastructure.

Global disparities are even more pronounced. Only 5\% of Africa's AI research community has access to the computational power that is needed for complex AI tasks, while 95\% rely on limited free tools~\cite{tsado_lee_2024_africa}. Compared with the 30-minute cycles that are available in G7 countries, African researchers face six-day iteration cycles, creating fundamental research capability inequities that no amount of algorithmic innovation can overcome.

\paragraph{Computational Infrastructure Misalignments}

Many researchers in the computational sciences are also members of the high-performance computing (HPC) community, which has long operated under a different set of constraints and practices. HPC systems are generally optimized for large sequential reads and writes (e.g., simulation checkpoints), not the millions of tiny files that are typically present in modern ML datasets~\cite{Lewis_2025}. For instance, system administrators may be displeased to train models directly on ImageNet-style datasets where every image is a small, individual file, as this can place enormous strain on the metadata servers of the file system by forcing constant indexing. This divergence in infrastructure expectations can create additional friction: ML researchers may be surprised when practices that are routine in their community are discouraged in some HPC settings, which can complicate collaboration on shared computing resources or with shared tooling.

\section*{Solutions and Implementation Strategies}

If these social and technical obstacles were meaningfully addressed, the scientific community could begin to unlock more of the long-term promise of AI, although not without continued risk, investment, and coordination. For example, researchers from the biology, chemistry, physics, and materials science fields might more systematically pool data and methods to construct shared, cross-domain models rather than working in isolated silos~\cite{neurips2024foundation}. Laboratories of varying sizes could deploy tighter experiment--model feedback loops that compress cycles from weeks to days, as demonstrated by autonomous laboratories that have successfully synthesized dozens of new materials through continuous operations~\cite{szymkuc2024insilico}. With better standards for provenance, metadata, and reproducibility, trusted AI assistants could help scientists spot anomalies, suggest hypotheses, and contextualize findings, while humans would remain central to the interpretation and validation processes. Moreover, communities could rally around common benchmarks and shared ``grand challenge'' problems to coordinate collective progress across different disciplines. Importantly, more equitable infrastructure, whether open datasets, low-cost computing platforms, or regional capacity-building, could broaden participation, making these capabilities accessible to researchers in underresourced institutions or geographies.

\subsection*{Solution One: Strengthening Collaboration and Education Across Communities}

\paragraph{Standardized Interfaces and APIs}

Technical collaboration barriers can be addressed through standardized interfaces that abstract implementation details while preserving domain-specific requirements. Machine learning libraries should provide scientific computing interfaces that automatically handle common preprocessing steps, domain-specific evaluation metrics, and uncertainty quantification ~\cite{wilkinson2016fair, ong2013python, polykovskiy2020molecular}.

Successful examples include scikit-image for microscopy~\cite{vanderwalt2014scikit} and astropy for astronomical data~\cite{astropy2013}, which demonstrate how standardized interfaces can expose simple methods to domain scientists while handling complex domain requirements such as irregular sampling conditions, missing data, and measurement uncertainties. Version control and reproducibility tools that are adapted for scientific workflows can facilitate collaboration by tracking code and data provenance~\cite{jacobsen2020fair}, although tools such as DVC and MLflow need extensions to handle domain-specific requirements such as experimental metadata and regulatory compliance~\cite{regev2017human}.

\paragraph{Collaborative Development Platforms and Training}
Successful technical collaboration requires platforms that enable iterative development between domain experts and ML researchers. While platforms such as Hugging Face demonstrate an effective community-driven development paradigm, scientific applications require additional features, including uncertainty quantification capabilities, domain-specific evaluation metrics, and validations against established scientific benchmarks~\cite{chanussot2021open}. Building scientific AI repositories that incorporate these requirements by default would substantially lower the barriers to collaboration~\cite{lhoest2021datasets}.

This requires training programs that build hybrid expertise rather than attempting to turn domain scientists into ML specialists or vice versa~\cite{ade2019simons}. The community should develop specialized roles for scientific AI practitioners who can effectively bridge both domains~\cite{karniadakis2021physics}. Online educational resources should prioritize practical skills for scientific AI applications--preparing domain datasets, validating ML models against scientific knowledge, and interpreting results in context~\cite{linkert2010metadata}. These resources should emphasize understanding over automation, enabling researchers to use AI tools responsibly rather than as opaque black boxes.

\paragraph{Case Study: Interdisciplinary Training Models}

The Eric and Wendy Schmidt AI in Science Postdoctoral Fellowship, a program of Schmidt Sciences, demonstrates effective approaches for building interdisciplinary capacity. The program annually supports approximately 160 postdoctoral fellows across nine universities globally~\cite{schmidtsciences2024aiinscience}. Fellows receive dual mentorship opportunities combining domain science expertise with AI technical skills, recognizing that effectively using AI for science requires specialized training rather than superficial collaboration between separate teams. The program structures cohort-based learning schemes where fellows develop a shared vocabulary across different disciplines, provides access to computational resources, and creates explicit career pathways that reward bridge-building contributions. Early outcomes suggest that compared with ad hoc partnerships, dedicated interdisciplinary roles, when properly supported at an institutional level, can more effectively overcome collaboration barriers between isolated research groups.

\subsection*{Solution Two: Structuring Upstream Challenges and Shared Benchmarks}

\paragraph{Community Mechanisms for Identifying Problems}

The research community needs systematic approaches for identifying computational bottlenecks with broad applicability through dedicated conference sessions and cross-disciplinary workshops. Several machine learning communities have already begun addressing data and benchmark gaps. For example, the NeurIPS Datasets and Benchmarks Track has established expectations concerning documentation, evaluation protocols, and dataset governance. However, these efforts remain largely internal to the ML research ecosystem and are not yet integrated with the domain-specific validation pipelines, experimental constraints, and long-term data stewardship that are required for scientific discovery.

Meaningful AI-for-science benchmarks require sustained
coordination between domain scientists, experimental facilities, and
ML researchers, not only standardized ML evaluation formats. To that end, prize competitions have demonstrated effective models for achieving community-driven problem identification: the Vesuvius Challenge~\cite{vesuvius_challenge} and DREAM Challenges~\cite{iscb2024_dream} have engaged more than 30,000 participants across sixty competitions, producing 105+ publications, while the DrivenData competitions~\cite{drivendata2024} have provided \$4.8M+ in prizes across 245,000 submissions advancing climate, health, and education solutions. These competitions attract diverse expertise to problems that individual research groups cannot address alone, using hybrid competition--collaboration models that prevent knowledge hoarding while maintaining competitive motivation.

Research funding programs should explicitly prioritize computational challenges with broad applicability rather than funding numerous domain-specific applications, as concentrated efforts on upstream problems can achieve a greater total impact. The community should develop frameworks for assessing the potential downstream impacts of computational advances to guide the strategic allocation of resources.

\paragraph{Building a Reusable Computational Infrastructure}

Successful upstream solutions require careful attention to generalizability and reusability through open-source development practices, well-documented APIs, comprehensive test suites, and modular architectures that enable researchers to build upon and extend existing computational tools~\cite{eisty2025essentialguidelinesbuildinghighquality}. The community should prioritize sustainability through proper software engineering practices rather than treating research codes as disposable prototypes while ensuring that reproducible science is valued and rewarded in terms of hiring, tenure evaluations, and citation counts.

Benchmark datasets and evaluation metrics for upstream problems should focus on computational efficiency, the accuracy attained across diverse problem instances, and uncertainty quantification rather than the performance achieved in narrow tasks~\cite{weidinger2025toward}. Recent work has advocated for comprehensive benchmark suites rather than single leaderboard scores, allowing for a better understanding of the tradeoffs between AI models without hiding the potential harms within aggregate metrics~\cite{wang2024benchmarksuites}. For the scientific workforce to coalesce around such challenges, domain leaders must take responsibility for defining them and actively engaging their communities via easy-to-use open benchmark creation methods~\cite{shashidhar2025yourbench}. The most effective step that individual researchers can take is to help articulate and participate in community benchmarks that frame upstream computational challenges as shared problems rather than siloed pursuits.

\paragraph{Case Study: The CASP and Upstream Problem Structuring}

The critical assessment of protein structure prediction (CASP) exemplifies how the coordination of a community around upstream computational challenges enables transformative breakthroughs. Launched in 1994 by John Moult and Krzysztof Fidelis, the CASP has established biennial blind prediction competitions with standardized evaluation protocols~\cite{moult1995caspexperiment}. Participating groups submit structure predictions before the experimental structures are released, ensuring that objective assessments are free from confirmation bias.

The success of the CASP stems from its social infrastructure as much as its technical design. The organizers built institutional norms around open participation, rigorous evaluation, and shared problem definition schemes that have persisted across decades and leadership transitions. The target selection protocols ensured that the predictions remained truly blind, whereas multiple assessment methods provided consensus validations. The competition format created both collaborative knowledge-sharing (through postassessment workshops and publications) and competitive motivation frameworks for advancing methods. Sustained NIH funding from 1994 to 2025 enabled continuity, although recent funding challenges underscore the fragility of such infrastructure when government support wavers.

This groundwork proved critical for the breakthrough of AlphaFold at CASP14 in 2020~\cite{jumper2021alphafold}. The technical innovation of DeepMind succeeded precisely because the CASP had established clear success metrics, assembled challenging test cases, and created a community that was ready to validate and build upon advances. Without the two decades of problem structuring and community coordination provided by the CASP, even sophisticated neural architectures would have struggled to demonstrate meaningful progress in the protein structure prediction domain. This case illustrates how upstream problem structuring through sustained community investment can multiply downstream impacts compared to those provided by fragmented efforts involving narrow applications.

\subsection*{Solution Three: Standardizing and Curating Scientific Data for Broad Reuse}

\paragraph{Community-Driven Data Standardization}

The successful democratization of scientific data requires practical, widely adopted standards that balance usability with scientific integrity. Simple, universally compatible formats such as CSV or Parquet often lower the barriers to collaboration more effectively than complex, domain-specific schemas do. Early bioinformatics efforts, such as MAGE-ML \cite{spellman2002_mage_ml}, illustrated the pitfalls of overly complex standards. Despite the technical flexibility of this method, its complexity hindered its interoperability, leading the community to adopt the simpler spreadsheet-based MAGE-TAB format \cite{rayner2006_magetab, rayner2009_magetabulator}. As noted earlier regarding the Cancer Imaging Archive \cite{clark2013cancer}, the lack of standardized schemas necessitates hundreds of hours of manual labor--effort that could be obviated by adopting simple, consistent formats.

The Protein Data Bank (PDB) exemplifies a large-scale, community-driven data democratization scheme. Starting with seven structures in 1971, it now features more than 230,000 three-dimensional protein structures with 10 million daily downloads~\cite{berman2000protein}. Its success stems from its use of standardized formats (PDB/mmCIF), expert curation, open access principles, and sustained funding. Similar approaches--minimal metadata standards, conversion tools for legacy formats, and validation pipelines--can enable other domains to achieve scalable reuse and collaborative development~\cite{pmid24109559, wilkinson2016fair, grossman2016toward}. The FAIR principles (Findable, Accessible, Interoperable, and Reusable) are being increasingly applied not only to data but also to AI models and software, although their implementation remains challenging and requires community consensus~\cite{ravi2024fairai}. Tools such as AIDRIN (AI Data Readiness Inspector) now provide comprehensive frameworks for assessing the readiness of data, including FAIR compliance, quality, understandability, value, fairness, and bias metrics~\cite{aidrin2024}. The 2021-2024 FAIR Data Spaces project in Germany laid the foundation for developing common data spaces from organizational, legal, and technical perspectives~\cite{fraunhofer2024fair}.

Even the modest adoption of open, accessible formats can have an outsized impact. Open-source tools such as Pandas, Polars, and Apache Arrow provide robust cross-platform support for scientific data processing, facilitating more than a billion computations daily \cite{virtanen2020scipy}. The growth of repositories such as Zenodo, which hosts more than 2 million datasets following the simple Dublin Core metadata standards, underscores how accessible standards accelerate the data sharing and democratization processes \cite{zenodo}.

\paragraph{Developing Architectures for Scientific Data}

High-dimensional, low-sample-size problems require architectural approaches that extend beyond the current transformer models. Progress has been made in terms of developing architectures that are tailored for scientific data. By explicitly modeling spatial relationships, graph neural networks have shown promise for use in molecular and material applications, achieving 20\% improvements in catalyst activity prediction tasks over the traditional descriptors \cite{gilmer2017neuralmessagepassingquantum, gasteiger2022directionalmessagepassingmolecular, Chen_2019}. Hierarchical attention mechanisms have been proposed to address the long-range interaction problem encountered in biological systems, enabling the analysis of gigapixel pathology images with spatial hierarchies spanning six orders of magnitude \cite{cao2021swinunetunetlikepuretransformer, chen2022scalingvisiontransformersgigapixel, Han_2023}. Physics-informed neural networks incorporate domain knowledge directly into their model designs, reducing the induced simulation errors by 2-3 orders of magnitude in fluid dynamics applications while requiring 100x less training data \cite{bdcc6040140, wang2020understandingmitigatinggradientpathologies, Kovacs_2022}. State-space models such as Mamba \cite{gu2023mamba} offer promising alternatives for handling the long sequences that are common in scientific time series, achieving linear scaling with the sequence length and contrasting with the quadratic complexity inherent in transformers. However, architectures still need better inductive biases concerning data structures and relationships, and effectively embedding such assumptions remains nontrivial~\cite{vafa2025foundationmodelfoundusing, Lake2017}. Deep knowledge of both machine learning and the relevant domain is essential for achieving success. Furthermore, each of the research areas mentioned above lacks robust benchmarking and consequently has struggled to gain sustained traction \cite{bechlerspeicher2025positiongraphlearninglose}.

However, architectural innovation alone is insufficient without addressing the sample efficiency problem. Few-shot learning approaches, meta-learning techniques, and transfer learning from simulation data offer promising directions, with recent demonstrations showing that 90\% accuracy is achievable with only 50 experimental samples when a model is pretrained on 10 million simulated materials \cite{Minami2025, baalouch2019simtorealdomainadaptationhigh, ohana2025welllargescalecollectiondiverse, wang2020understandingmitigatinggradientpathologies, Alsentzer2025}. The community should prioritize the development of foundation models that capture general scientific principles and can be fine-tuned for specific applications with limited data, which requires coordinated dataset construction efforts spanning multiple institutions and decades of sustained funding commitment.

\paragraph{Case Study: The Materials Project}

The Materials Project has demonstrated successful large-scale data democratization in computational materials science. Launched in 2011 with continuous NSF and DOE funding, it now provides open access to computed properties for more than 154,000 known and predicted materials~\cite{jain2013materials, persson2025accelerated}. Its impact stems from deliberate attention to usability alongside technical infrastructure.

The Materials Project's broad adoption beyond well-resourced institutions reflects deliberate design choices prioritizing accessibility. Its MongoDB backend uses JSON-based data formats with well-documented RESTful APIs rather than domain-specific complexity~\cite{ong2013python}. Extensive documentation and tutorials serve both computational researchers and experimentalists. Clear data provenance and versioning enable users to trace computational parameters and reproduce results, while governance mechanisms incorporate community feedback into curation priorities.

The Materials Project now serves more than 600,000 registered users globally, with sustained institutional investments spanning over a decade, thus demonstrating that infrastructure democratization requires long-term commitment rather than short-term grants~\cite{persson2025accelerated}. Its success illustrates that when a data infrastructure receives appropriate community governance and continuous support, researchers at underresourced institutions can participate meaningfully in computational discoveries.

\subsection*{Solution Four: Building Accessible and Sustainable Infrastructure}

\paragraph{Efficient Model Sharing and Deployment Schemes}

Democratization requires an infrastructure for sharing not only trained models but also entire scientific AI pipelines. Scientific models often require complex preprocessing, specialized evaluation, and careful uncertainty quantification processes that generic model repositories do not adequately support.

Model efficiency becomes critical for providing equitable access. Quantization, pruning, and knowledge distillation can dramatically reduce the imposed computational requirements while preserving scientific accuracy~\cite{dantas2024comprehensive}, but these techniques need adaptations for scientific applications rather than simply borrowing consumer-focused optimizations~\cite{marino2023deep}. For instance, scientific models must maintain calibrated uncertainty estimates even after undergoing compression. Edge deployment strategies are particularly important for field research, remote laboratories, and institutions with limited computational infrastructure~\cite{fahim2021hls4ml}.

\paragraph{Community-Owned Infrastructure and Sustainable Funding}

Sustainable infrastructure depends on community ownership rather than a reliance on commercial platforms or individual institutions. Open-source projects such as Jupyter, Conda, and the Scientific Python ecosystem illustrate successful community-driven models that serve broad research needs. Provenance tracking records the origins and transformations of datasets and analyses, which is essential for conducting reproducible research~\cite{jacobsen2020fair}.

Building this infrastructure is critical, but so is investment in outreach and dissemination. Valuable resources already exist---such as free computing, storage, and API credits from NAIRR~\cite{nsf_nairr_2024}, OSPool~\cite{osg_pool}, and Hugging Face~\cite{huggingface_storage_limits_2025}---yet often go unused because potential users remain unaware of them. Without active education and visibility, infrastructure risks being underutilized.

The central challenge is the acquisition of sustainable funding that supports both long-term development and the communication strategies that are needed for adoption. Infrastructure takes years of effort to build, but traditional research funding schemes favor novelty over maintenance~\cite{crawford2024nasa}. Addressing this gap will require mechanisms that combine support for infrastructure and dissemination through consortium models, government investments, or hybrid public--private partnerships~\cite{chanzuckerberg2024eoss}.

\paragraph{Case Study: Open-Source and Federated Infrastructure}

Several initiatives demonstrate the appetite among scientists for community-owned infrastructure that enables collaborative research. The Hugging Face Hub provides free hosting for more than 2 million models and 500,000 datasets serving 11 million users with built-in version control, standardized metadata, and collaborative features~\cite{wolf2020huggingface}. Researchers share artifacts with simple commands, discover relevant work through searches, and reproduce results through integrated tools. The platform's recent extensions, with the help of community code contributions, to support domain-specific scientific data formats including FASTA files for genomic sequences, CIF files for crystallographic structures, and streaming protocols for large HDF5 physics datasets, reflect community demand for infrastructure that bridges machine learning tools with scientific data practices. The Hugging Science initiative attracted 7,000 Discord members and 800 Hub contributors within months of launching, with events generating waitlists in the thousands. The Open Science Grid similarly demonstrates sustained community engagement through a consortium model with NSF and DOE support since 2004, providing US-based researchers with access to distributed computing resources across participating institutions~\cite{pordes2007opensciencegrid, sfiligoi2009osg}. Its federated architecture allows institutions to contribute resources while maintaining local control, and the platform's consistent growth for 10 years indicates genuine research community need for such shared infrastructure.

However, building accessible platforms proves easier than sustaining the communities they enable. The labor required to maintain these initiatives, including curating datasets, writing documentation, moderating discussions, providing technical support, establishing governance structures, and ensuring quality standards, falls almost entirely outside the reward mechanisms that govern academic careers. Grant applications rarely fund infrastructure maintenance. Tenure committees do not recognize community moderation. Citation metrics cannot capture the value of helping hundreds of researchers debug their workflows. This misalignment creates fragility: even successful platforms depend on individuals willing to perform essential but unrecognized work, or on external funding commitments vulnerable to shifting institutional priorities. The Open Science Grid's stability relies on continued government support rather than inherent sustainability. Hugging Science's rapid growth required substantial volunteer effort with uncertain long-term viability. Infrastructure democratization therefore requires more than technical accessibility. It demands that scientific institutions fundamentally restructure how they value, recognize, and compensate the collaborative work that makes shared research infrastructure possible. Without such changes, platforms may proliferate while the communities they aim to serve remain unable to sustain them.

\section*{Discussion}

AI systems already provide meaningful value to scientists by accelerating literature synthesis, assisting with data workflows, and helping researchers navigate rapidly expanding knowledge domains~\cite{fda2023aiml,merchant2023gnome}. Yet these successes also reveal something crucial: there remains a substantial gap between automating scientific tasks and genuinely advancing scientific discovery. The risk is that narratives celebrating ``autonomous scientists'' distract us from addressing the social and structural barriers scientists face.

As we document throughout this paper, these obstacles continue to persist because of long time institutional practices that shape scientific work. Laboratories continue using legacy formats because they lack incentives or institutional support to adopt new standards~\cite{bhardwaj2024mlcuration}. Benchmark fragmentation continues because no neutral organizations exist with stable funding to steward collective resources~\cite{wang2024benchmarksuites}. Infrastructural inequity reflects decades of concentrated investment in privileged institutions~\cite{ahmed2020dedemocratization,tsado_lee_2024_africa}, while essential labor like data curation remains systematically undervalued~\cite{fecher2015reputationeconomyresultsempirical}. Interdisciplinary collaboration struggles because institutional incentives reward narrow methodological innovation over the patient work of bridging fields.

The transformative potential of AI for science therefore depends less on building increasingly autonomous models than on strengthening the sociotechnical systems in which those tools operate. Progress requires durable, community-governed benchmarks and practical data standards, infrastructure aligned with scientific workflows, and training pathways that genuinely bridge disciplines. Realizing this vision means confronting structural causes: realigning academic incentives, securing stable funding, and building governance mechanisms robust enough to outlast individual projects. Progress will come from coordinated communities sustaining shared resources and collective stewardship, not from models acting in isolation.

\paragraph{Limitations}

This work synthesizes the challenges encountered across a wide range of scientific domains--biology, chemistry, materials science, physics, and climate science--to highlight the observed structural patterns. While these themes are broadly shared, the specific constraints and opportunities vary by field. As a result, some recommendations may need to be adapted for particular subdisciplines or regulatory environments. In addition, several of the arguments rely on studies and surveys that do not encompass the full diversity of scientific institutions, especially in underrepresented regions. The rapid pace of progress that is being achieved in machine learning also means that some technical limitations discussed here may evolve, even if the underlying social and institutional dynamics persist. Finally, our discussion of governance and incentives is necessarily a high-level process; implementing these ideas would require detailed policy designs and negotiations among stakeholders.

\paragraph{Future Work}

Future research should investigate how these sociotechnical barriers manifest within specific research communities to generate actionable guidance that is sensitive to local norms and constraints. Conducting comparative studies of successful and unsuccessful attempts to establish data standards, shared benchmarks, or community-governed infrastructure would clarify which governance models best support long-term coordination~\cite{ravi2024fairai,aidrin2024}. On the technical side, research into models that incorporate mechanistic priors, integrate simulation and experimental feedback loops, and provide principled uncertainty quantification capabilities may help align AI systems more closely with scientific reasoning~\cite{Minami2025,ohana2025welllargescalecollectiondiverse}. Global-scale studies of computational access and capacity building are also needed to ensure that the progress attained in AI for science benefits a broad scientific community rather than deepening the existing disparities~\cite{tsado_lee_2024_africa}.

\paragraph{Declaration of Interest}
Both authors are employed by Hugging Face, which is referenced in this work as one provider of collaborative platforms and infrastructure for AI research. To address this potential conflict, our analysis encompasses the broader ecosystem of service providers and infrastructure models, including academic, government, and commercial initiatives beyond Hugging Face. Our recommendations focus on community-wide standards and practices rather than promoting specific platforms. The systemic barriers and solutions we identify apply regardless of which service providers researchers choose to use. We have strived to maintain objectivity by grounding our analysis in peer-reviewed literature, publicly documented case studies, and challenges documented across multiple institutions and platforms.



\newpage

\bibliographystyle{plainnat}
{\small
\bibliography{ref}
}

\end{document}